\documentclass{article} 
\usepackage[preprint]{colm2026_conference}

\usepackage{microtype}
\usepackage{hyperref}
\usepackage{url}
\usepackage{booktabs}
\usepackage{graphicx}
\usepackage[most]{tcolorbox}
\usepackage{xcolor}         
\usepackage{xspace}
\usepackage{array}
\usepackage{multirow}
\usepackage{enumitem}

\usepackage{lineno}

\definecolor{darkblue}{rgb}{0, 0, 0.5}
\hypersetup{colorlinks=true, citecolor=darkblue, linkcolor=darkblue, urlcolor=darkblue}

\title{AgentIR: Reasoning-Aware Retrieval for Deep Research Agents}

\author{%
Zijian Chen$^1$,
Xueguang Ma$^1$,
Shengyao Zhuang$^2$, \\
\textbf{Jimmy Lin$^1$},
\textbf{Akari Asai$^3$},
\textbf{Victor Zhong$^1$}\\
[1ex] 
$^1$University of Waterloo,\quad
$^2$University of Queensland,\quad
$^3$Carnegie Mellon University\\
[1.5ex]
}

\newcommand{\method}{\texttt{\textsc{Reasoning-Aware Retrieval}}\xspace}
\newcommand{\model}{\texttt{\textsc{AgentIR-4B}}\xspace}
\newcommand{\modelsynth}{\texttt{\textsc{DR-Synth}}\xspace}

\usepackage{etoolbox}
\newtoggle{comment}
\settoggle{comment}{false}  

\begin{document}

\ifcolmsubmission
\linenumbers
\fi

\maketitle
\lhead{Preprint. Work In Progress.}
\thispagestyle{fancy}

\begin{abstract}
Deep Research agents are rapidly emerging as primary consumers of modern retrieval systems. Unlike human users who issue queries without documenting their intermediate thought processes, Deep Research agents generate explicit natural language reasoning before each search, revealing rich intent and context that existing retrievers entirely ignore. To exploit this signal, we introduce: (1) \textbf{\method}, a retrieval paradigm that jointly embeds the agent's reasoning trace alongside its query; and (2) \textbf{\modelsynth}, a data synthesis method that generates Deep Research retriever training data from standard QA datasets. We demonstrate that both components are independently effective, and their combination yields a trained embedding model, \model, with substantial gains. On the challenging BrowseComp-Plus benchmark, \model achieves 68\% accuracy with the open-weight agent Tongyi-DeepResearch, compared to 52\% with conventional embedding models twice its size, and 37\% with BM25. Code and data are available at: \url{https://texttron.github.io/AgentIR/}. 
\end{abstract}

\begin{figure*}[h]
    \centering
    \includegraphics[width=\linewidth]{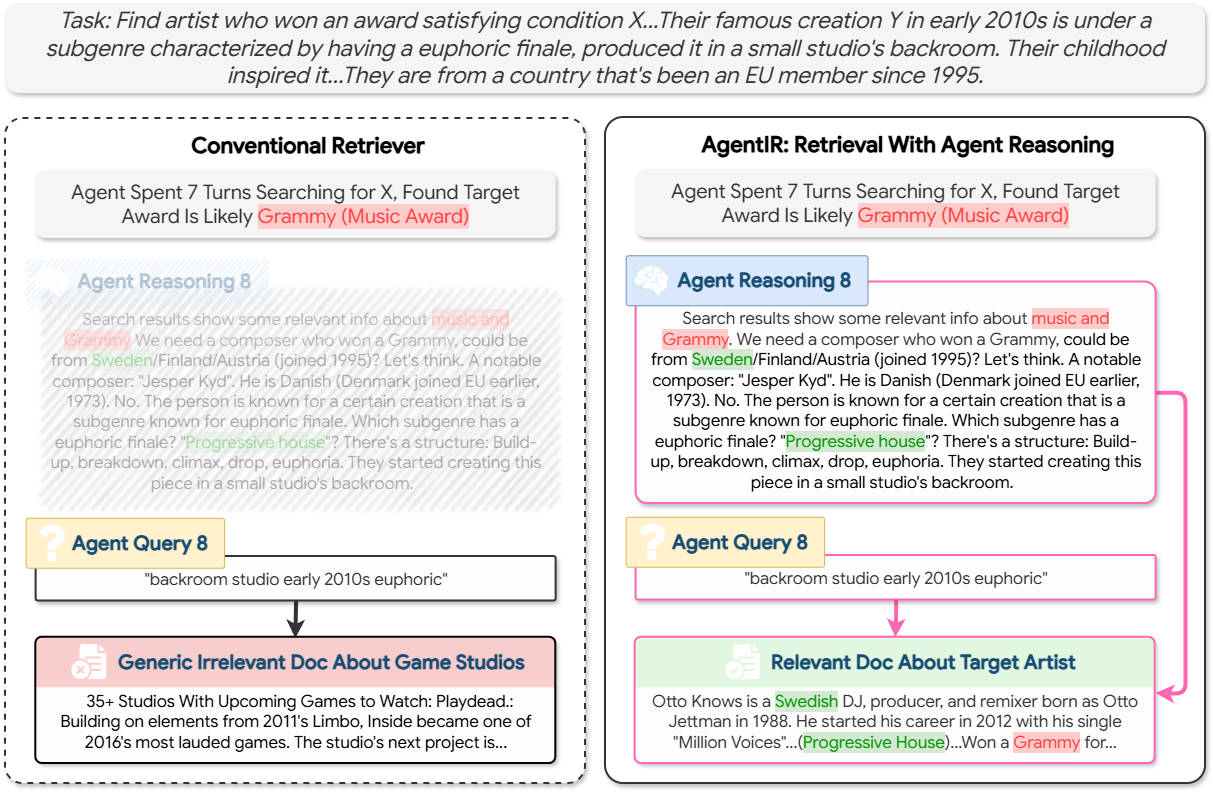}
    \caption{\method (\model) vs.\ conventional retrieval (Qwen3-Embedding-4B) for a task from BrowseComp-Plus, paired with the Tongyi-DR agent. The task has been simplified for display.}
    \label{fig:teaser}
\end{figure*}

\section{Introduction}

Deep Research agents, large language models (LLMs) that autonomously reason and search across multiple turns, have emerged as a new class of users of retrieval systems~\citep{advancing_search_with_agents,browsecomp,webarena, tongyi-dr, search-r1, websailor, webshaper}. Unlike human users who issue and refine queries without documenting their intermediate thought processes, Deep Research agents generate explicit, natural language reasonings before every search call. These reasoning traces encode rich signals about search intent and the evolving problem-solving context. Yet, no existing retriever learns to exploit them. 

Consider the example in Figure~\ref{fig:teaser}. At turn 8 of a multi-turn search process, the agent issues the query ``backroom studio early 2010s euphoric''. Conventional retrieval with this ambiguous query alone yields generic, irrelevant results. However, the agent's preceding reasoning trace reveals the broader objective: to find a composer who won an award $X$, and composed music $Y$ in the 2010s in a small studio's backroom, within a subgenre known for ``euphoric finale''. In fact, it indicates that the previous searches have already identified award $X$ as the ``Grammy''. Further, drawing from its parametric knowledge, the agent hypothesizes that a subgenre with a ``euphoric finale'' is likely ``progressive house'', which turns out to be correct. Here, the reasoning trace provides highly informative signals: reflection on prior results, identification of unresolved gaps, and hypotheses about promising search targets.

We propose \method, a new retrieval paradigm that exploits this observation: instead of embedding only the agent's issued query, we jointly embed the reasoning trace, learning a retriever that leverages the rich intent and contextual information expressed in agent reasoning. Further, to address the lack of retriever training data for agent-issued sub-queries in Deep Research, we introduce \modelsynth, a data synthesis method that transforms standard QA datasets such as WebShaper~\citep{webshaper} into (agent sub-query, relevance) pairs tailored for Deep Research agent retrieval.

Training \method on synthesized data derived from WebShaper yields \model, an embedding model that substantially outperforms prior retrievers on BrowseComp-Plus~\citep{bcp, browsecomp}, a challenging Deep Research benchmark. Paired with the open-weight agent Tongyi-DeepResearch (Tongyi-DR)~\citep{tongyi-dr}, \model achieves 68\% end-to-end accuracy, compared to 52\% for a strong conventional embedding model twice its size, and 37\% for BM25. It also outperforms computationally intensive methods such as LLM-based reranking by 10\% absolute.

Beyond accuracy, \model improves efficiency by reducing the number of search steps required to complete tasks, and does not require additional inference overhead compared to prior query-rewriting methods, since the reasoning traces it leverages are already generated ``for free''. 

Importantly, without additional training, these gains generalize across other agents with different reasoning patterns, such as gpt-oss-120B and GLM-4.7.

Further analysis of alternative retrieval signals shows that reasoning traces are effective not only because they summarize relevant findings from earlier turns, but also because they implicitly filter out outdated or incorrect information, yielding a cleaner  signal for retrieval.

In summary, our contributions are:
\begin{itemize}
  \item We propose \method, a retrieval paradigm for Deep Research agents that leverages agent reasoning traces to improve retrieval.
  \item We propose \modelsynth, a data synthesis method that constructs (agent sub-query, relevance) pairs from standard QA datasets, addressing the lack of training data in retrieval for Deep Research agents.
  \item We train \model, achieving an 18\% absolute accuracy gain over strong conventional retrievers on BrowseComp-Plus, generalizable across different agent models without additional training.
\end{itemize}

\section{Related Work}

\paragraph{Deep Research Agents.} Recently, Retrieval-Augmented Generation (RAG)~\citep{rag,rag2} has evolved from single-turn retrieve-then-answer pipelines to language models that autonomously conduct multi-turn searches through test-time scaling to solve complex problems~\citep{selfrag}. This paradigm has been further accelerated by reinforcement learning, leading to a new generation of ``Deep Research agents''~\citep{tongyi-dr,search-r1,websailor, webshaper}. These agents perform extensive exploration, often executing over 20 retrieval turns to resolve tasks that would take humans hours to complete~\citep{browsecomp, bcp}. Compared to prior retrievers designed for single-turn RAG, Deep Research's inherent multi-turn nature presents a new retrieval problem, which \method aims to address.

\paragraph{Retrieval and Reasoning.} A notable feature of Deep Research agents is their ability to interleave explicit reasoning and retrieval in solving complex tasks, which we learn to leverage. In parallel, retrievers such as ReasonIR~\citep{reasonir} and RaDeR~\citep{rader} have emerged to tackle reasoning-intensive tasks without an agent. We emphasize that \method addresses a fundamentally different objective: rather than expecting the retriever to solve a complex task in a single turn, we focus on collaborative retrieval with an agent to resolve complex tasks over multiple turns.

\paragraph{Understanding Ambiguous Queries.} Understanding and handling ambiguous human user queries has been a long-standing challenge in information retrieval~\citep{ambiguous_queries,estimating_query_difficulty,predicting_query_performance}. Instruction-aware retrieval addresses this by incorporating explicit human-written instructions~\citep{tart}. In interactive settings, systems may also ask clarifying questions to disambiguate the user's intent~\citep{clarifying_question}. When no human annotation is available, methods such as HyDE~\citep{hyde} prompt an LLM to interpret an ambiguous query, and enrich it with hypothetical context from parametric knowledge. All of these approaches share a common premise: the query itself is an inherently under-specified representation of the user's true intent, and therefore requires mining additional signal. In Deep Research, this signal is available for free: the agent's reasoning trace explicitly articulates its intent, and \method learns to exploit it.

\section{Methodology}

\subsection{Preliminary}

\paragraph{Deep Research.}
Following prior work~\citep{tongyi-dr}, we formulate Deep Research as a ReAct-style~\citep{react} loop, in which an LLM agent interacts with a retriever over multiple turns to solve complex tasks.
The agent's behaviour can be represented as a trajectory of $(\tau_i, a_i, o_i)$ turns:
\begin{align*}
  \mathcal{H}_T = (\tau_1, a_1, o_1, \cdots, \tau_i, a_i, o_i, \cdots, \tau_T, a_T)
\end{align*}
At each turn $t \leq T$, the LLM's policy $\pi$ generates a reasoning trace $\tau_t$ and an action $a_t$ conditioned on the interaction history: $\tau_t, a_t \sim \pi(\cdot \mid \mathcal{H}_{t-1})$.
The agent then receives feedback $o_t$ from the environment.
In this work, unless otherwise specified, an action $a_t$ is either a search call issued by the agent to the retriever (yielding results $o_t$), or a final answer that terminates the loop.
For simplicity, we use $q_t$ and $a_t$ interchangeably when the action is a search call.

\paragraph{Conventional Retrieval in Deep Research.}
Existing approaches treat retrieval for a Deep Research agent's query identically to a standalone human search: given query $q_t$, the retriever $R$ searches using only this query ($o_t \gets R(q_t)$). 
A fundamental limitation of this setup is that a query alone is often under-specified with respect to the user's underlying intent, posing ambiguity and persistent challenges for retrieval systems~\citep{ambiguous_queries,estimating_query_difficulty}. 
To mitigate this, query expansion methods such as HyDE~\citep{hyde} prompt an LLM to enrich ambiguous queries with hypothetical relevant content; effectively, when the user's underlying thought process is unavailable, an external LLM's interpretation is used as a proxy.

\subsection{Reasoning-Aware Retrieval} \label{sec:method}
Conventional retrieval underutilizes the structure of Deep Research agents. Unlike human users, Deep Research agents explicitly
expose the reasoning traces that motivate their search queries, and this transparent reasoning process should be exploited by retrievers.
To this end, we propose \method, a paradigm that jointly embeds the reasoning trace alongside the query ($o_t \gets R(\tau_t, q_t)$), using a concatenation template shown in Figure~\ref{fig:prompt_main}.

To illustrate the value of reasoning traces in action, consider the example in Figure~\ref{fig:teaser}. The trace enhances retrieval in three ways:
\begin{itemize}
  \item \textbf{Task Intent:} The query ``backroom studio early 2010s euphoric'' is inherently ambiguous. However, the reasoning trace clarifies its intent: ``finding a composer who composed euphoric music in the early 2010s in a backroom studio''. Without this reasoning, a conventional retriever misinterprets the query as a search for video game studios. Analogous to human-written instructions in task-aware retrieval~\citep{tart}, the reasoning trace here functions as an implicit agent-written instruction.
  \item \textbf{Reflection on Prior Results:} Unique to Deep Research is its multi-turn nature, which requires incorporating prior results. In this example, the overall task is to find an artist who won a specific award $X$ and composed a specific song $Y$. As reflected in the reasoning trace, previous searches have already identified award $X$ as ``Grammy,'' drastically narrowing the search space for the target artist.
  \item \textbf{Hypothetical Search Targets:} Beyond incorporating known information from past results, the agent uses its parametric knowledge to \emph{infer} likely targets for future searches. In Figure~\ref{fig:teaser}, the agent hypothesizes that a country ``that's been an EU member since 1995'' is likely ``Sweden/Finland/Austria'', and that the music subgenre with a ``euphoric finale'' is likely ``progressive house.'' Both hypotheses are correct, further narrowing the search space. This behaviour may resemble HyDE. However, importantly, HyDE enriches the query using parametric knowledge alone, unaware of any agent state; in contrast, the agent's reasoning is generated using parametric knowledge \emph{and} the full interaction history: $\tau_t \sim \pi(\cdot \mid \mathcal{H}_{t-1})$, yielding hypotheses that are far more grounded in the agent's evolving context.
\end{itemize}

Importantly, unlike task-aware retrieval~\citep{tart} that requires explicit human instructions, or HyDE that necessitates an additional, costly LLM call purely for query expansion, the Deep Research agent generates its reasoning trace entirely ``for free'' as part of its standard operating loop.

\begin{figure*}[t]
    \centering
    \includegraphics[width=\linewidth]{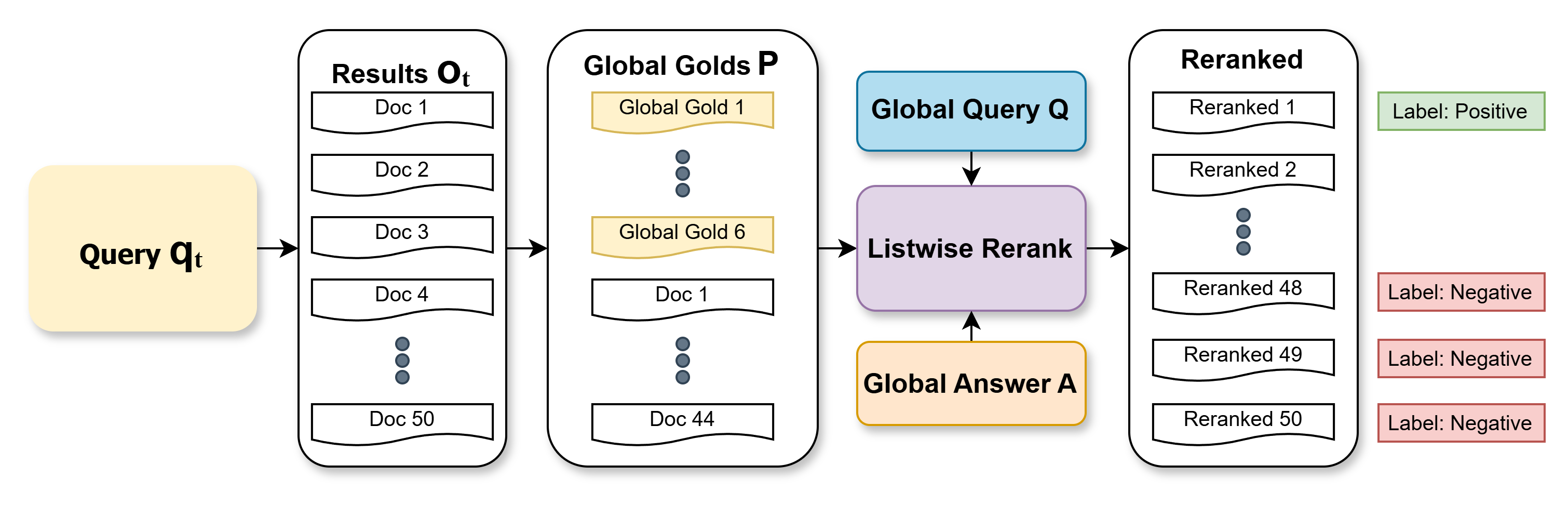}
    \caption{Oracle reranking procedure used in \modelsynth (Section~\ref{sec:labels})}
    \label{fig:oracle_rerank}
\end{figure*}

\subsection{\modelsynth: Constructing Training Data for Deep Research Queries} \label{sec:labels}

While the embedding models powering modern retrieval are increasingly capable, they are explicitly pre-trained on query-document pairs rather than reasoning-heavy agent traces~\citep{qwen3-embed}.
Consequently, off-the-shelf retrievers are not optimized to appropriately weight the reasoning traces and queries embedded.
To bridge this gap and execute \method effectively, we must train the retriever to align reasoning-augmented queries with relevant documents.
We achieve this using a standard contrastive learning loss~\citep{contrastive_learning}:
\begin{align}
    -\log \frac{\exp\left(\operatorname{sim}([\tau_t, q_t], d^+_t)/T\right)}{\exp\left(\operatorname{sim}([\tau_t, q_t], d^+_t)/T\right) + \sum_{d^-_t \in \{d^-_t\}} \exp\left(\operatorname{sim}([\tau_t, q_t], d^-_t)/T\right)} \label{eq:contrastive}
\end{align}
Here, $[\tau_t, q_t]$ denotes the concatenation of the agent reasoning $\tau_t$ and query $q_t$, $d^+_t$ is a positive document, $\{d^-_t\}$ is a set of negative documents, $\operatorname{sim}$ denotes cosine similarity, and $T = 0.01$ is the temperature used.

\paragraph{The Need for New Training Data.} However, this poses a challenge: there is currently no retriever training data tailored to sub-queries $q_t$'s in multi-turn Deep Research. Traditional QA and single-turn information retrieval datasets provide $(Q, A, P)$ triples consisting of a global question $Q$, an answer $A$, and a set of positive documents $P$ sufficient to answer $Q$~\citep{msmarco}. Yet, in multi-turn Deep Research, the agent observes the global $Q$ and iteratively issues local sub-queries $q_t$'s; the retriever's task is to handle these local queries $q_t$'s, not the original $Q$. Further, we lack relevance supervision $d^+_t, \{d^-_t\}$ for these local sub-queries: the positive documents $P$ provided by existing datasets serve the global $Q$, whereas each local sub-query $q_t$ typically targets only a subset of the global $Q$.

To address this gap, we propose \modelsynth, a data synthesis pipeline that generates sub-query level retriever training data from standard QA datasets by leveraging agent rollouts.

\paragraph{Generating Sub-Queries.}
Given an agent, a conventional query-only retriever, and a standard dataset of $(Q, A, P)$ triples, we construct sub-queries as follows. For each $(Q, A, P)$, we run the agent with the query-only retriever on $Q$, producing a trajectory $\mathcal{H}_T$ of $T$ turns: $T-1$ search turns followed by a final answer turn. From this trajectory, we extract the reasoning-query pairs $(\tau_t, q_t)$ at each search turn $t$, for $1 \leq t \leq T - 1$. Consequently, each original question $Q$ yields $T - 1$ turns of sub-query instances for training.

\paragraph{Generating Supervision.}
To provide relevance labels for a sub-query at turn $t$, we must identify documents relevant to the specific clues sought at that turn. Further, unlike single-turn retrieval, Deep Research requires explicitly considering the global objective in $Q$: for instance, if $Q$ seeks an entity satisfying three conditions $X, Y, Z$, and the current turn $t$ targets only $X$, a document satisfying $X$ but violating $Y$ and $Z$ should be ranked lower than another document satisfying all three. 

To generate labels that are both relevant to turn $t$ and aligned with $Q$, we modify the rollout process described above. At each retrieval turn $t$, instead of directly passing the retrieval results to the next turn, we refine the candidate ranking with an oracle reranking procedure, and derive labels from the refined ranking:

\begin{enumerate}
  \item Retrieve the top 50 documents using the conventional query-only retriever.
  \item Prepend the positive documents $P$ to the candidate list. These documents are guaranteed to be useful for the global question $Q$ overall. Since the current turn addresses a subset of the reasoning hops, it is likely that some candidates from $P$ are highly relevant to the current turn $t$.
  \item Prompt an LLM to perform listwise reranking~\citep{rankgpt, lrl} over the candidate pool. We prompt the LLM with the current query $q_t$, the global question $Q$, and $Q$'s true answer $A$, instructing it to rank the documents based on their relevance to $q_t$ while ensuring alignment with the overall $(Q, A)$ pair. The full prompt can be found in Figure~\ref{fig:prompt_oracle_rerank}.
  \item Label the top-ranked document as the positive document $d^+_t$ for turn $t$, and the bottom seven documents as hard negatives $\{d^-_t\}$. Figure~\ref{fig:oracle_rerank} visualizes this process.
\end{enumerate}

After processing a single $(Q, A, P)$ triple, we obtain $T-1$ training instances of $([\tau_t, q_t], d^+_t, \{d^-_t\})$. Following prior work~\citep{tongyi-dr}, after performing rollouts for all $(Q, A, P)$ triples in the dataset, we apply rejection sampling and train only on rollouts that successfully answered $Q$.

\section{Experiments}

\subsection{Training Setup}

We instantiate \method with \modelsynth-generated data, training a concrete model, \model. Specifically, we apply \modelsynth to WebShaper~\citep{webshaper} $(Q, A, P)$ triples, producing 5,238 training instances of $([\tau_t, q_t], d^+_t, \{d^-_t\})$. These instances are used to fine-tune Qwen3-Embedding-4B~\citep{qwen3-embed} with contrastive learning.
During rollout generation, we use Tongyi-DeepResearch (Tongyi-DR)~\citep{tongyi-dr} as the agent and Qwen3-Embedding-8B~\citep{qwen3-embed} as the query-only retriever. Additional details on data construction and model training are provided in Appendix~\ref{appendix:training}.

\subsection{Evaluation Setup and Metrics}

To assess retrievers in end-to-end Deep Research, we pair them with three open-weight Deep Research agents: Tongyi-DeepResearch (Tongyi-DR)~\citep{tongyi-dr}, gpt-oss-120B (oss-120b-high),\footnote{\href{https://openai.com/index/introducing-gpt-oss/}{https://openai.com/index/introducing-gpt-oss/}, high reasoning effort} and GLM-4.7,\footnote{\href{https://z.ai/blog/glm-4.7}{https://z.ai/blog/glm-4.7}} evaluating them on BrowseComp-Plus~\citep{bcp}, a benchmark featuring complex multi-hop queries requiring 20+ searches. Following its official evaluation, we give each agent a ``search'' tool that retrieves top-5 document snippets truncated to 512 tokens. Tongyi-DR is also trained to use a ``visit'' tool that opens a full document in addition to the ``search'' tool. For fair comparison, we record its accuracy both with and without this tool. 

For each agent-retriever combination, we report end-to-end QA {\bf Accuracy} following the same LLM-as-judge setup as BrowseComp-Plus, the {\bf Recall} of all documents ever retrieved by the agent's search calls against the ground-truth evidence documents, and the number of {\bf Search Calls} the agent issued before giving a final answer. 

\subsection{Baselines}

\paragraph{Query-Only Retrievers. } We compare against strong conventional retrievers that use the query only: the Qwen3-Embedding-4B backbone before fine-tuning, the classic sparse retriever BM25~\citep{bm25}, a strong dense retriever Qwen3-Embedding-8B~\citep{qwen3-embed}, and a reasoning-intensive retriever ReasonIR-8B~\citep{reasonir}.\footnote{We reran the BrowseComp-Plus baseline for oss-120b-high with an updated vllm version, which attained higher accuracy than in the original paper.}

\paragraph{Query Expansion.} As discussed in Section~\ref{sec:method}, \method's method of embedding agent reasoning traces relates to past query expansion methods like HyDE. To this end, we compare with  Reason-Rewriter\footnote{\href{https://huggingface.co/cfli/reasoner-rewriter-qwen2.5-7b-0821}{https://huggingface.co/cfli/reasoner-rewriter-qwen2.5-7b-0821}} +  Reason-Embed-8B~\citep{reasonembed},\footnote{\href{https://huggingface.co/hanhainebula/reason-embed-qwen3-8b-0928}{https://huggingface.co/hanhainebula/reason-embed-qwen3-8b-0928}} a fine-tuned HyDE-style query expander, paired with its dedicated retriever, shown to perform well on reasoning-intensive tasks. Further, we compare with Agentic-R~\citep{agenticr}, a concurrent work that also trains retrievers specialized for Deep Research agents, where they expand the agent's query $q_t$ by prepending the global query $Q$, following the notation in Section~\ref{sec:labels}.

\paragraph{Reranking.} To provide a strong reference, we also evaluate listwise reranking~\citep{rankgpt}, a computationally expensive reranking method, applied to the first-stage Qwen3-Embedding-4B retriever. Specifically, we use Qwen3-8B~\citep{qwen3} to rerank the top-20 retrieved documents.

\begin{table}[t]
  \centering
  \small
  \setlength{\tabcolsep}{6pt}
  \renewcommand{\arraystretch}{1.15}

  \begin{tabular}{l l r r c}
    \toprule
    \textbf{LLM} & \textbf{Retriever}                & \textbf{Accuracy} & \textbf{Recall} & \textbf{Search Calls}       \\
    \midrule

    \multirow{8}{*}{Tongyi-DR}
                 & BM25                              & 33.98             & 46.83           & 32.92                       \\
                 & Qwen3-Embed-4B                    & 48.67             & 59.90           & 31.02                       \\
                 & Qwen3-Embed-8B                    & 50.72             & 61.78           & 30.43                       \\
                 & ReasonIR-8B                       & 51.03             & 63.62           & 31.15                       \\
                 & Reason-Rewriter + Reason-Embed-8B & 31.08             & 40.15           & 34.64                       \\
                 & Agentic-R                         & 44.70             & 47.67           & 31.05                       \\
                 & Qwen3-Embed-4B + LLM Rerank       & 55.66             & 68.35           & 28.85                       \\
    \cmidrule{2-5}
                 & \model                            & \textbf{66.27}    & \textbf{78.86}  & 25.91                       \\
    \midrule

    \multirow{8}{*}{oss-120b-high}
                 & BM25                              & 36.02             & 43.32           & 31.00                       \\
                 & Qwen3-Embed-4B                    & 47.59             & 58.15           & 29.14                       \\
                 & Qwen3-Embed-8B                    & 49.52             & 60.70           & 28.74                       \\
                 & ReasonIR-8B                       & 50.84             & 60.71           & 29.03                       \\
                 & Reason-Rewriter + Reason-Embed-8B & 32.21             & 38.51           & 33.17                       \\
                 & Agentic-R                         & 45.66             & 46.42           & 28.53                       \\
                 & Qwen3-Embed-4B + LLM Rerank       & 53.49             & 64.55           & 27.41                       \\
    \cmidrule{2-5}
                 & \model                            & \textbf{66.99}    & \textbf{78.13}  & 24.08                       \\
    \midrule

    \multirow{8}{*}{GLM-4.7}
                 & BM25                              & 33.25             & 45.97           & 41.55                       \\
                 & Qwen3-Embed-4B                    & 50.48             & 66.05           & 35.38                       \\
                 & Qwen3-Embed-8B                    & 50.18             & 68.69           & 35.32                       \\
                 & ReasonIR-8B                       & 52.27             & 68.22           & 35.94                       \\
                 & Reason-Rewriter + Reason-Embed-8B & 34.90             & 49.68           & 41.13                       \\
                 & Agentic-R                         & 46.75             & 52.09           & 35.20                       \\
                 & Qwen3-Embed-4B + LLM Rerank       & 55.54             & 73.19           & 34.27                       \\
    \cmidrule{2-5}
                 & \model                            & \textbf{64.66}    & \textbf{79.21}  & 29.85                       \\
    \midrule

    \multirow{8}{*}{Tongyi-DR (visit)}
                 & BM25                              & 36.87             & 43.02           & 30.73 + 2.75 Visit          \\
                 & Qwen3-Embed-4B                    & 50.24             & 58.42           & 29.45 + 3.14 Visit          \\
                 & Qwen3-Embed-8B                    & 51.93             & 60.51           & 29.31 + 3.11 Visit          \\
                 & ReasonIR-8B                       & 52.65             & 61.49           & 29.68 + 3.13 Visit          \\
                 & Reason-Rewriter + Reason-Embed-8B & 29.76             & 36.65           & 32.65 + 2.62 Visit          \\
                 & Agentic-R                         & 45.54             & 46.39           & 30.88 + 2.83 Visit          \\
                 & Qwen3-Embed-4B + LLM Rerank       & 54.35             & 65.22           & 28.04 + 3.05 Visit          \\
    \cmidrule{2-5}
                 & \model                            & \textbf{68.07}    & \textbf{76.58}  & 24.49 + 3.41 Visit          \\
    \bottomrule
  \end{tabular}
  \vspace{-0.5em}
  \caption{End-to-end evaluation on BrowseComp-Plus. For Tongyi-DR with the visit tool, Search Calls are reported as (search + visit). LLM-Rerank refers to listwise reranking the top-20 results with Qwen3-8B.}
  \label{tab:main}
\end{table}

\subsection{Results} \label{sec:end-to-end}

Table~\ref{tab:main} reports end-to-end Deep Research results on BrowseComp-Plus. \model achieves the best performance across all agents, substantially outperforming all prior baselines and competitive concurrent work.

For Tongyi-DR, \model achieves 66.27\% accuracy, a 17.60\% absolute improvement over the Qwen3-Embed-4B backbone. To put this in perspective, this gain is comparable to the accuracy gain from BM25 to Qwen3-Embed-4B (14.69\%). Additionally \model also outperforms Qwen3-Embed-8B, a model twice its size, by $\approx 15$\% absolute accuracy.

Beyond accuracy, \model achieves notable efficiency gains: the number of search calls decreases from 32.92 with BM25 to 25.91 with \model. Moreover, \model outperforms Qwen3-Embed-4B + LLM Rerank, a highly computationally expensive reranking method by approximately 10\% absolute accuracy, despite performing no reranking.

\paragraph{\model Outperforms Reasoning-Intensive Retrievers and Query Expansion.} \model surpasses past single-turn reasoning-intensive retrievers such as ReasonIR-8B by $\approx 15$\% accuracy. Further, HyDE-like query expanders, such as Reason-Rewriter + Reason-Embed-8B, appear ineffective in Deep Research settings. Without access to agent context, we found that their hypotheses often lead to misinterpretations and substantial hallucinations: as shown in Figure~\ref{fig:hyde_example_full}, given the query ``backroom studio early 2010s euphoric'', the expander misinterprets the intended ``backroom of a studio'' as a studio named ``Backroom Studio'', with hallucinated ``Los Angeles-based'' and ``Social Media Impact''. In contrast, \model's direct access to agent reasoning yields much more grounded hypotheses, as shown in Figure~\ref{fig:teaser}.

\paragraph{\model Generalizes Across Agents.} \model demonstrates strong generalization across multiple dimensions. First, it is trained on WebShaper queries, making the improvements on BrowseComp-Plus zero-shot. Second, although Tongyi-DR is used to generate training trajectories, Table~\ref{tab:main} shows that \model transfers effectively to gpt-oss-120B and GLM-4.7, agents with distinct reasoning styles and search behaviours, without any additional fine-tuning. Finally, \model can be applied in combination with other tools: when used alongside the visit tool in the Tongyi-DR (Visit) setting, we observe an equal level of gains.

\section{Analysis}

We now analyze the source of $\model$'s effectiveness by ablating its two core components, the use of agent reasoning traces and training on synthetic data. Further, we study whether alternative signals beyond the agent reasoning can also improve retrieval, and analyze why they may fall short of \model.

\begin{table}[h]
  \centering
  \small
  \setlength{\tabcolsep}{5pt}
  \renewcommand{\arraystretch}{1.15}

  \begin{tabular}{l l r r c}
    \toprule
    \textbf{Agent} & \textbf{Method}                     & \textbf{Accuracy} & \textbf{Recall} & \textbf{Search Calls} \\
    \midrule

    \multirow{4}{*}{Tongyi-DR}
                   & Qwen3-Embed-4B                                & 48.67             & 59.90           & 31.02                 \\
                   & \model (w/o Training)                        & 55.54             & 66.13           & 29.21                 \\
                   & \model (w/o Reasoning)                       & 59.40             & 70.02           & 27.97                 \\
                   & \model                                       & \textbf{66.27}    & \textbf{78.86}  & 25.91        \\
    \midrule

    \multirow{4}{*}{oss-120b-high}
                   & Qwen3-Embed-4B                                & 47.59             & 58.15           & 29.14                 \\
                   & \model (w/o Training)                        & 51.33             & 63.05           & 27.72                 \\
                   & \model (w/o Reasoning)                       & 59.16             & 68.80           & 26.64                 \\
                   & \model                                       & \textbf{66.99}    & \textbf{78.13}  & 24.08        \\
    \midrule

    \multirow{4}{*}{GLM-4.7}
                   & Qwen3-Embed-4B                                & 50.48             & 66.05           & 35.38                 \\
                   & \model (w/o Training)                        & 50.90             & 65.88           & 34.04                 \\
                   & \model (w/o Reasoning)                       & 57.47             & 75.07           & 32.92                 \\
                   & \model                                       & \textbf{64.66}    & \textbf{79.21}  & 29.85        \\
    \midrule

    \multirow{4}{*}{Tongyi-DR (visit)}
                   & Qwen3-Embed-4B                                & 50.24             & 58.42           & 29.45 + 3.14 Visit    \\
                   & \model (w/o Training)                        & 53.98             & 61.55           & 28.09 + 3.13 Visit    \\
                   & \model (w/o Reasoning)                       & 59.52             & 67.05           & 26.60 + 3.10 Visit    \\
                   & \model                                       & \textbf{68.07}    & \textbf{76.58}  & 24.49 + 3.41 Visit \\
    \bottomrule
  \end{tabular}
  \vspace{-0.5em}
  \caption{Component ablation. All methods use Qwen3-Embed-4B as the backbone. ``\model (w/o Training)'' prepends reasoning traces without additional fine-tuning. ``\model (w/o Reasoning)'' trains on \modelsynth-generated WebShaper data (Section~\ref{sec:labels}) but embeds only the query.}
  \label{tab:ablation_components}
\end{table}

\subsection{Effectiveness of \method and \modelsynth}

While Table~\ref{tab:main} shows that the full \model system outperforms all baselines, how much do the \method paradigm and training on the synthetic data from \modelsynth contribute individually? To disentangle their effects, we evaluate each component independently on top of the Qwen3-Embed-4B backbone: \textbf{\model (w/o Training)} jointly embeds the agent's reasoning trace and query using the frozen backbone embedding model; \textbf{\model (w/o Reasoning)} trains the backbone using our \modelsynth-generated data, but embeds only the agent's query during training and inference.

As shown in Table~\ref{tab:ablation_components}, both components are independently effective, and their combination delivers the largest gains. On Tongyi-DR, jointly embedding the agent reasoning without any fine-tuning improves accuracy from 48.67\% to 55.54\%. Training the retriever without using reasonings achieves 59.4\%, and when combined with reasonings, accuracy further increases to 66.27\%. This pattern holds consistently across all three agents.

\subsection{Alternative Sources of Retrieval Signals} \label{sec:ablation-variants}

Beyond the current reasoning and query, we investigate whether other components of the trajectory $\mathcal{H}_t$ can also serve as effective retrieval signals, and how they compare to \model.

Formally, at turn $t$ with trajectory $\mathcal{H}_t = (\tau_1, q_1, o_1, \cdots, \tau_t, q_t)$, \model extracts $f(\mathcal{H}_t) = (\tau_t, q_t)$ as retrieval input. We study alternative choices of $f$, where each alternative is trained as a separate retriever following the same setup as \model. Specifically:
\begin{itemize}
    \item \textbf{Prior Queries}: $f(\mathcal{H}_t) = (q_1, q_2, \cdots, q_t)$. Prior work in conversational search~\citep{convdr} shows that embedding prior user queries improves retrieval. We evaluate how this approach compares to using reasoning traces.
    \item \textbf{Prior Queries \& Reasonings}: $f(\mathcal{H}_t) = (\tau_1, q_1, \cdots, \tau_t, q_t)$. Building on the previous setting, we test whether augmenting prior queries with their corresponding reasoning traces yields additional gains.
    \item \textbf{Prior Queries \& Reasonings \& Docs}: $f(\mathcal{H}_t) = \mathcal{H}_t$. As a stress test, we embed the full trajectory, including all prior reasonings, queries, and retrieved documents. However, due to context length limitations, we truncate to only use the most recent turns, averaging around 3 turns for the agents studied.
    \item \textbf{Global Question}: $f(\mathcal{H}_t) = (Q, q_t)$. Beyond the trajectory $\mathcal{H}_t$, we also evaluate explicitly including the global question $Q$, following notation from Section~\ref{sec:labels}. This setting is motivated by concurrent work Agentic-R~\citep{agenticr}.
\end{itemize}

\paragraph{Setup.}
Each alternative is trained as a separate retriever following the same setup as \model, with the only difference being the input construction: instead of extracting $(\tau_t, q_t)$ from $\mathcal{H}_t$ during sub-query generation, we use each corresponding $f(\mathcal{H}_t)$ defined above. Additionally, we also report a ``None'' setting, corresponding to the \model (w/o Reasoning) entry in Table~\ref{tab:ablation_components}, where $f(\mathcal{H}_t) = q_t$. More implementation details can be found in Appendix~\ref{appendix:variants}.

\paragraph{Global Question.}
Concurrent work Agentic-R utilizes the Global Question setting, shown ineffective in Table~\ref{tab:main}. However, their experiments use a different backbone (E5~\citep{e5}), and are trained on HotPotQA~\citep{hotpotqa} and TriviaQA~\citep{triviaqa}, datasets that contain only $\approx 2$ hops per question. To isolate whether the signal $Q$ itself benefits retrieval, we replicate a checkpoint using the same backbone and data as \model.

\begin{table}[t]
  \centering
  \small
  \setlength{\tabcolsep}{5pt}
  \renewcommand{\arraystretch}{1.15}

  \begin{tabular}{l l r r c}
    \toprule
    \textbf{Agent} & \textbf{$f(\mathcal{H}_t)$} & \textbf{Accuracy} & \textbf{Recall} & \textbf{Search Calls} \\
    \midrule

    \multirow{6}{*}{Tongyi-DR}
                   & None                                & 59.40             & 70.02           & 27.97                 \\
                   & Current Reasoning ({\bf \model})    & \textbf{66.27}    & \textbf{78.86}  & 25.91                 \\
                   & Global Question                              & 63.25             & 70.98           & 26.90                 \\
                   & Prior Queries                       & 63.13             & 71.55           & 27.12                 \\
                   & Prior Queries \& Reasonings         & 63.13             & 74.20           & 26.48                 \\
                   & Prior Queries \& Reasonings \& Docs & 60.00             & 70.73           & 27.18                 \\
    \midrule

    \multirow{6}{*}{oss-120b-high}
                   & None                                & 59.16             & 68.80           & 26.64                 \\
                   & Current Reasoning ({\bf \model})    & \textbf{66.99}    & \textbf{78.13}  & 24.08                 \\
                   & Global Question                              & 61.93             & 70.10           & 25.66                 \\
                   & Prior Queries                       & 61.89             & 71.53           & 25.41                 \\
                   & Prior Queries \& Reasonings         & 64.34             & 73.32           & 24.46                 \\
                   & Prior Queries \& Reasonings \& Docs & 58.67             & 67.80           & 25.55                 \\
    \midrule

    \multirow{6}{*}{GLM-4.7}
                   & None                                & 57.47             & 75.07           & 32.92                 \\
                   & Current Reasoning ({\bf \model})    & \textbf{64.66}    & \textbf{79.21}  & 29.85                 \\
                   & Global Question                              & 61.20             & 73.54           & 29.77                 \\
                   & Prior Queries                       & 59.08             & 74.96           & 31.11                 \\
                   & Prior Queries \& Reasonings         & 60.80             & 75.20           & 29.60                 \\
                   & Prior Queries \& Reasonings \& Docs & 58.67             & 70.84           & 30.18                 \\
    \midrule

    \multirow{6}{*}{Tongyi-DR (visit)}
                   & None                                & 59.52             & 67.05           & 26.60 + 3.07 Visit    \\
                   & Current Reasoning ({\bf \model})    & \textbf{68.07}    & \textbf{76.58}  & 24.49 + 3.41 Visit    \\
                   & Global Question                              & 63.73             & 68.25           & 25.33 + 3.33 Visit    \\
                   & Prior Queries                       & 63.01             & 69.91           & 26.44 + 3.17 Visit    \\
                   & Prior Queries \& Reasonings         & 66.27             & 74.25           & 25.22 + 3.17 Visit    \\
                   & Prior Queries \& Reasonings \& Docs & 61.45             & 67.32           & 25.61 + 3.37 Visit    \\
    \bottomrule
  \end{tabular}
  \vspace{-0.5em}
  \caption{Ablation over alternative signals. All models are fine-tuned from Qwen3-Embedding-4B using \modelsynth (Section~\ref{sec:labels}), varying only the trajectory transformation $f(\mathcal{H}_t)$ used as the retriever's input. ``None'' embeds only the query, corresponding to the \model (w/o Reasoning) entry in Table~\ref{tab:ablation_components}. }
  \label{tab:ablation}
\end{table}

\paragraph{Results.}
Table~\ref{tab:ablation} shows consistent trends across all agents. \model outperforms Prior Queries and Global Question, confirming that reasoning traces provide a distinct and valuable retrieval signal. Augmenting Prior Queries with their associated Reasonings improves over Prior Queries alone, but still underperforms \model.

The Prior Queries \& Reasonings \& Docs setting yields only modest improvements over ``None'' for Tongyi-DR and GLM-4.7, with a slight degradation for gpt-oss-120B. This behaviour likely stems from noise introduced by retrieved documents: irrelevant searches are augmented to subsequent queries, propagating further errors.
Indeed, for Tongyi-DR under the Prior Queries \& Reasonings \& Docs setting, 11.45\% of the runs had zero recall, meaning no evident document is retrieved at any turn. These runs average 37.46 search turns, compared to 27.18 overall, indicating prolonged and compounding retrieval failures driven by noisy context.

\subsection{Effect of Adding Prior Reasonings} \label{sec:num-turns}

The previous section demonstrates that reasoning traces provide a strong retrieval signal. Yet, incorporating the full history of prior reasonings does not improve over using just the current turn’s reasoning. We investigate this trend more closely: does adding more prior reasoning beyond the current turn actually improve retrieval?

To study this, we train additional checkpoints following the same setup, while varying the number of prior turns included in the input. Rather than just the most recent turn, we embed the past $k \in \{1, 2, 5, 9, 17, \text{all}\}$ turns. Formally, at turn $t$, we use $f(\mathcal{H}_t) = (\tau_j, q_j, \cdots, \tau_t, q_t)$ where $j = \max(1, t-k+1)$. Under this definition, $k=1$ corresponds to \model.

\begin{figure*}[t]
    \centering
    \includegraphics[width=\linewidth]{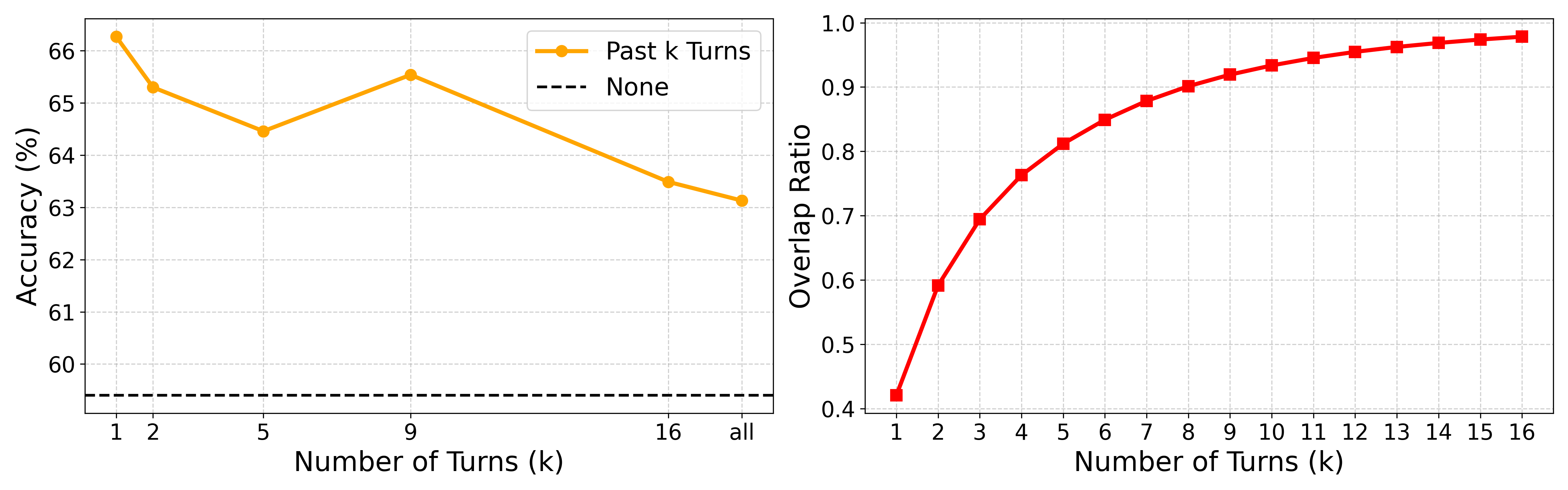}
    \caption{Effect of embedding $k$ history turns. We fix the Agent to Tongyi-DR. Plot (a) shows the end-to-end accuracy of embedding past $k$ turns, where ``None'' denotes the ``\model w/o Reasoning'' entry in Table~\ref{tab:ablation_components}. Plot (b) shows the ratio of unique clues covered by using $k$ most recent reasonings, among all clues that have been covered. This is averaged across all trajectories for the $k=\text{all}$ setting (Section~\ref{sec:num-turns}). }
    \label{fig:ablation_turns}
\end{figure*}

\begin{figure}[t]
    \centering
    \begin{minipage}[c]{0.5\linewidth}
        \fcolorbox{black!30}{gray!5}{%
            \begin{minipage}{0.92\linewidth}
                \vspace{0.3em}
            \small
            Thus, we have identified the person: Otto Knows (real name Otto Jettman). The creation is ``Million Voices''. Let's ensure that this matches all clues:
            \begin{itemize}[leftmargin=*, noitemsep, topsep=2pt]
              \item From Sweden (EU member since 1995) -- yes.
              \item Won a Grammy in late 2010s -- In 2017, he won a Grammy for Best Dance Recording for ``I Need You''.
              \item Famous creation spawned early 2010s -- ``Million Voices'' released 2012.
              \item It is framed under subgenre characterized by euphoric finale -- progressive house often described as having uplifting euphoric drops. Could also be called ``progressive house'' which has euphoric climax.
              \item Childhood inspired it -- maybe he says the track is about his childhood? We need to verify. The Wikipedia snippet doesn't mention childhood inspiration. Let's search.
            \end{itemize}
            \end{minipage}%
        }
    \end{minipage}
    \hfill
    \begin{minipage}[c]{0.48\linewidth}
        \centering
        \includegraphics[width=\linewidth]{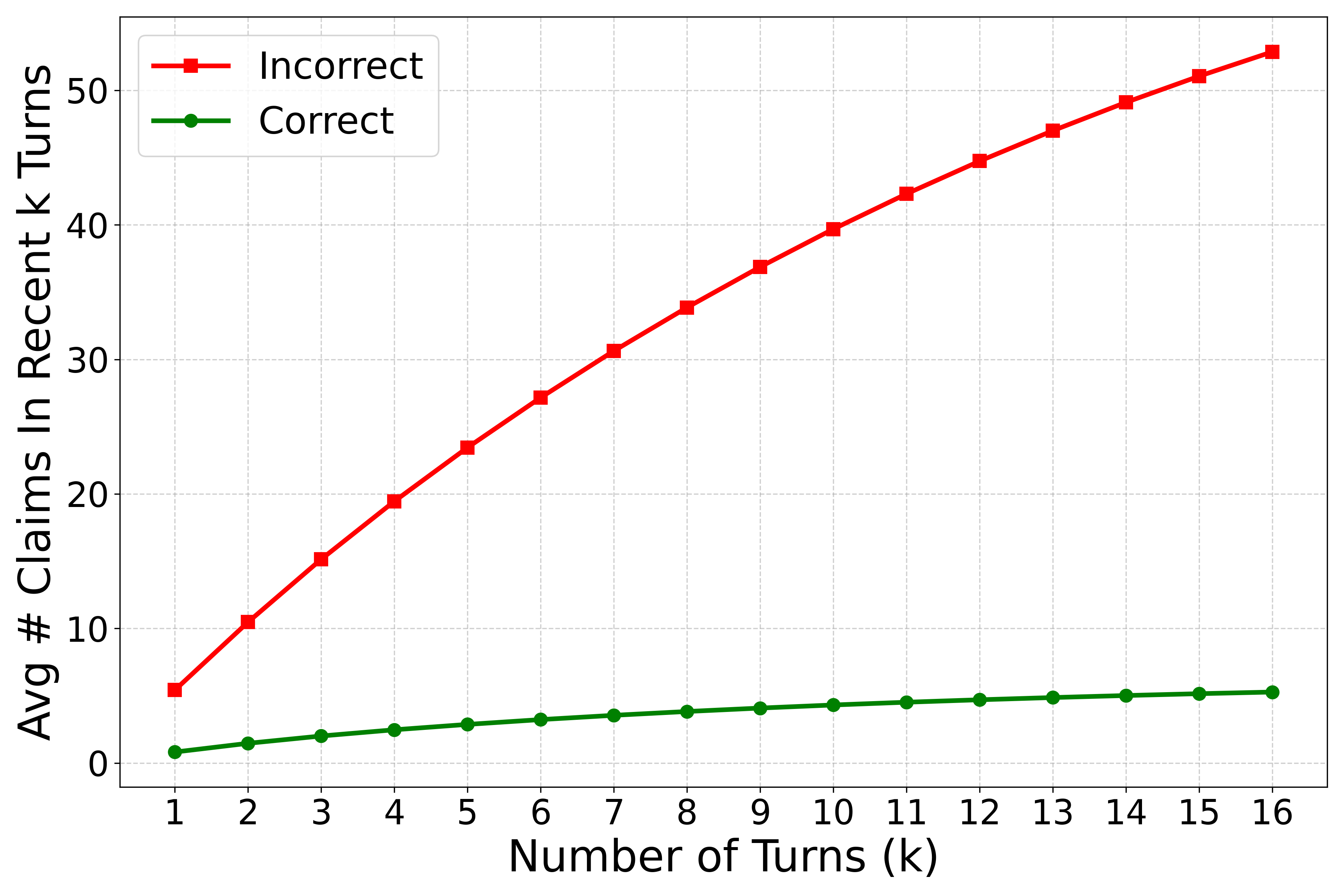}
    \end{minipage}
    \caption{(a) Reasoning for the query in Figure~\ref{fig:teaser} after identifying the candidate artist, Otto Knows from Sweden. (b) Average number of correct vs. incorrect claims (hypotheses) using $k$ most recent reasonings. This is averaged across trajectories for the $k$ = all setting (Section~\ref{sec:num-turns})}
    \label{fig:reasonings_overlap}
\end{figure}

Figure~\ref{fig:ablation_turns} (a) illustrates the end-to-end accuracy for different values of $k$ on Tongyi-DR. Overall, incorporating additional prior turns beyond the current reasoning does not improve accuracy. We next analyze potential causes for this behaviour.

\paragraph{Redundancy in Past Reasonings.} We hypothesize that the lack of improvement  is partly due to the repetition between reasonings across turns. Critically, even though we do not explicitly embed the full history, the current reasoning, $\tau_t \sim \pi(\cdot | \mathcal{H}_{t-1})$ is generated conditioned on the \emph{entire history}; consequently, it often summarizes findings from prior turns. Indeed, in our running example in Figure~\ref{fig:teaser}, after identifying the target artist and country, the agent's subsequent reasoning reiterates these key facts, as shown in Figure~\ref{fig:reasonings_overlap} (a).

To quantify this redundancy, we analyze trajectories from the $k=\text{all}$ setting. For each trajectory, we  prompt an LLM (GLM-4.7) with the full list of reasonings, and instruct it to decompose them into atomic clues. For each reasoning, we then ask the LLM to identify which atomic clues it contains, using the prompts in Appendix~\ref{appendix:atomic_clues}.

Based on these annotations, we compute a coverage metric: on average, what fraction of the clues from all history turns are covered by the $k$ most recent reasonings? Figure~\ref{fig:ablation_turns} (b) illustrates the results. The current reasoning alone ($k=1$) already covers more than 40\% of all past clues. As $k$ increases, coverage grows with clear diminishing returns, indicating that additional prior reasonings contribute little new information.

\paragraph{Forgetting as a Feature.} Beyond redundancy, earlier reasonings may actively introduce noise. Consider the example in Figure~\ref{fig:teaser}. When identifying ``a country that's been an EU member since 1995'', the agent brainstorms candidates such as ``Sweden/Finland/Austria''. Similarly, when searching for the artist, it hypothesizes ``Jesper Kyd''. While such speculative probing may be useful for the immediate next search, once the correct country and artist have been identified as ``Otto Knows from Sweden'', such incorrect hypotheses become noise for future retrieval.

Notably, notice that the reasoning's summary of prior results in Figure~\ref{fig:reasonings_overlap} (a) does not revisit failed candidates (e.g., Finland and Jesper Kyd). That is, the current reasoning not only makes additional prior turns redundant by summarizing confirmed findings, but also implicitly filters out outdated or incorrect guesses. We hypothesize that this implicitly curated past history provides a cleaner retrieval signal than naively embedding the entire reasoning history.

To measure the effect of noise, we prompt an LLM to extract both correct claims and incorrect claims or hypotheses from each reasoning. We then compute, on average, how many incorrect and correct claims appear within the most recent $k$ reasonings. As shown in Figure~\ref{fig:reasonings_overlap} (b), adding more prior turns introduces substantially more noise of incorrect claims than useful signal. More details about the noise measurement can be found in Appendix~\ref{appendix:noise_prompt}. 

\section{Conclusion}

We introduce \method, a paradigm designed for Deep Research agents. Unlike conventional retrieval systems that process isolated queries, we leverage the agent's explicit natural language reasoning traces, a rich contextual and intent signal absent in traditional search for humans. Additionally, to address the lack of multi-turn retriever training data for Deep Research, we propose \modelsynth, a  pipeline that synthesizes such data from standard QA datasets. Our experiments demonstrate that \method and \modelsynth are independently effective. Their combination, \model, substantially outperforms existing retrievers, with consistent gains across multiple LLM agents.

Our analysis reveals that \method is effective as the reasoning trace grounds each search in the agent's historical context. Further, it implicitly curates that history, where incorrect hypotheses from prior turns are naturally filtered out, and explicitly embedding more uncurated history underperforms. These findings motivate future work on ``context engineering'' for retrievers. While this term is typically used to describe managing agent context,\footnote{\href{https://www.anthropic.com/engineering/effective-context-engineering-for-ai-agents}{https://www.anthropic.com/engineering/effective-context-engineering-for-ai-agents}} we argue for its application to retrieval: developing principled curation methods that optimize the retriever's view of the evolving problem. What's more, as context engineering for agents advances, these agent-optimized contexts may be directly leveraged by retrievers, improving performance without additional computational cost.

As Deep Research agents become more capable and commercial products~\footnote{\href{https://gemini.google/overview/deep-research/}{https://gemini.google/overview/deep-research/}, \href{https://openai.com/index/introducing-deep-research/}{https://openai.com/index/introducing-deep-research/}} mature, we anticipate a shift in which humans increasingly delegate complex information seeking to autonomous agents; that is, agents become the primary consumers of search, and humans become consumers of agents. By releasing \model, we aim to encourage the information retrieval community to dedicate more research serving this emerging class of ``agent users''.

\bibliography{agentir}
\bibliographystyle{colm2026_conference}

\clearpage
\appendix
\section{Main \model Prompt} \label{appendix:prompt_main}

\begin{figure}[h]
    \centering
    \begin{tcolorbox}[
        enhanced,
        sharp corners,
        width=0.95\linewidth,
        colback=gray!5,
        colframe=gray!60,
        coltitle=white,
        colbacktitle=gray!60,
        fonttitle=\small\ttfamily\bfseries,
        title=Instruction:,
        boxrule=0.6pt,
        left=3mm,
        right=3mm,
        top=1mm,
        bottom=1mm,
        toptitle=0.3mm,
        bottomtitle=0.3mm,
        lefttitle=3mm,
        bottomrule=0pt,
        nobeforeafter,
        after skip=0pt
    ]
        \small\ttfamily
        Given a user's reasoning followed by a web search query, retrieve relevant passages that answer the query while incorporating the user's reasoning.
    \end{tcolorbox}
    \nointerlineskip\vspace{-0.6pt}
    \begin{tcolorbox}[
        enhanced,
        sharp corners,
        width=0.95\linewidth,
        colback=gray!5,
        colframe=gray!60,
        coltitle=white,
        colbacktitle=gray!60,
        fonttitle=\small\ttfamily\bfseries,
        title=Query:,
        boxrule=0.6pt,
        left=3mm,
        right=3mm,
        top=1mm,
        bottom=1mm,
        toptitle=0.3mm,
        bottomtitle=0.3mm,
        lefttitle=3mm,
        toprule=0.6pt,
        nobeforeafter,
        before skip=0pt
    ]
        \small\ttfamily
        Reasoning: \{reasoning\} \\
        Query: \{query\}
    \end{tcolorbox}
    \caption{The prompt template used to embed for \model. At turn $t$, we fill in \{reasoning\} with $\tau_t$ and \{query\} with $q_t$. Note that the duplicate ``Query:'' is intentional due to Qwen3-Embedding-4B~\cite{qwen3-embed}'s instruction format.}
    \label{fig:prompt_main}
\end{figure}

\section{\model Training Details} \label{appendix:training}

We adopt Qwen3-Embedding-4B~\citep{qwen3-embed} as the backbone model and fine-tune with LoRA~\citep{lora} on contrastive learning loss defined in Equation~\ref{eq:contrastive}. Each input consists of $[\tau_t, q_t]$, the concatenation of the reasoning trace $\tau_t$ and query $q_t$, formatted using the prompt in Figure~\ref{fig:prompt_main}. Training positives $d^+_t$ and hard negatives $\{d^-_t\}$ are generated by \modelsynth, together with standard in-batch negatives.

We fine-tune with a learning rate of 1e-4, batch size 4, maximum document length 4096, and maximum query length 8192. Training is conducted using the Tevatron~\citep{tevatron} toolkit on a single H100, with gradient checkpointing and gradient accumulation of 2, for 2 epochs on the \modelsynth-generated WebShaper dataset~\citep{webshaper}. 

\paragraph{Training Data Details.}
We apply \modelsynth to the 500 $(Q, A, P)$ pairs provided in WebShaper, performing rollouts using Tongyi-DR as the agent, Qwen3-Embedding-8B as retriever, and GLM-4.6 as oracle reranker, as detailed in Section~\ref{sec:labels}. At each turn of retrieval, we obtain the top 50 documents from the retriever and apply the reranking procedure. Then, we label the top-ranked document as positive $d^+_t$, the bottom-ranked seven as hard negatives $\{d^-_t\}$, and return the top 5 ranked documents back to the agent to continue its rollout. 

However, since WebShaper provides only the URLs of its positive documents $P$, without a complete document corpus, we construct a training corpus to ground WebShaper as follows:
\begin{itemize}
    \item \textbf{Positives.} For each positive document $p \in P$, we scrape its URL using Selenium,\footnote{\href{https://www.selenium.dev/documentation}{https://www.selenium.dev/documentation}} and parse the content using Trafilatura~\citep{trafilatura}.
    \item \textbf{Hard Negatives.} For each question $q \in Q$, we decompose it into $k$ sub-queries using GPT-4o, where $k$ is approximately 6 on average. Each sub-query is issued to a Google Search API provider, SerpAPI, which returns up to 100 results. We scrape these URLs using the same pipeline as for positives.
    \item \textbf{Random Negatives.} To better approximate realistic large scale retrieval, we add one million randomly sampled documents from FineWeb~\citep{fineweb} as random negatives.
\end{itemize}
After deduplication by URL, the training corpus consists of 1,146,942 documents. After rollout generation on this corpus, 250 of the 500 WebShaper queries were correctly answered. Using these 250 correct trajectories, we obtain a total of 5,238 training instances $([\tau_t, q_t], d^+_t, \{d^-_t\})$.

\begin{figure}[h]
    \centering
    \begin{tcolorbox}[
        enhanced,
        sharp corners,
        width=0.95\linewidth,
        colback=gray!5,
        colframe=gray!60,
        coltitle=white,
        colbacktitle=gray!60,
        fonttitle=\small\ttfamily\bfseries,
        title=System:,
        boxrule=0.6pt,
        left=3mm,
        right=3mm,
        top=1mm,
        bottom=1mm,
        toptitle=0.3mm,
        bottomtitle=0.3mm,
        lefttitle=3mm,
        bottomrule=0pt,
        nobeforeafter,
        after skip=0pt
    ]
        \small\ttfamily
        You are an intelligent assistant that can rank passages to best help an user in correctly answering a question. You will be given a complex question, and its correct answer. A user, who is trying to solve the complex question, issues a simpler probing web query, which retrieved some passages. You are to rank these passages based on their relevancy to the user's probing query, while prioritizing their usefulness in helping the user to correctly answer the complex question.
    \end{tcolorbox}
    \nointerlineskip\vspace{-0.6pt}
    \begin{tcolorbox}[
        enhanced,
        sharp corners,
        width=0.95\linewidth,
        colback=gray!5,
        colframe=gray!60,
        coltitle=white,
        colbacktitle=gray!60,
        fonttitle=\small\ttfamily\bfseries,
        title=User:,
        boxrule=0.6pt,
        left=3mm,
        right=3mm,
        top=1mm,
        bottom=1mm,
        toptitle=0.3mm,
        bottomtitle=0.3mm,
        lefttitle=3mm,
        toprule=0.6pt,
        nobeforeafter,
        before skip=0pt
    ]
        \small\ttfamily
        A user is trying to answer a complex question: \{question\}. The correct answer is: \{correct\_answer\}. A user, trying to solve the complex question, issued this probing web query: \{query\}, which retrieved \{num\} passages. I will provide you with these \{num\} passages, each indicated by a numerical identifier []. Rank the passages based on their relevance to the probing query, as well as their usefulness in helping the user to correctly answer the complex question. \\[0.5em]

        \relax[1]: \{doc 1\} \\
        \relax[2]: \{doc 2\} \\
        ... (repeat \{num\} times) \\[0.5em]

        Complex Question: \{question\}. \\
        Correct Answer: \{correct\_answer\}. \\
        Probing Query: \{query\}. \\
        Rank the \{num\} passages above based on their relevance to the probing query, as well as their usefulness in helping the user to correctly answer the complex question. All the passages should be included and listed using identifiers, in descending order of relevance. The output format should be [] > [], e.g., [2] > [1]. Answer concisely and directly and only respond with the ranking results, do not say any word or explain.
    \end{tcolorbox}
    \caption{Prompt template used for the oracle reranker (Section~\ref{sec:labels}).}
    \label{fig:prompt_oracle_rerank}
\end{figure}

\section{Prompts for Alternative Sources of Retrieval Signals} \label{appendix:variants}

For each variant $f(\mathcal{H}_t)$ described in Section~\ref{sec:ablation-variants}, we follow the same training procedure outlined in Appendix~\ref{appendix:training}. The only difference is the input text to retriever training. Instead of extracting the $(\tau_t, q_t)$ pairs from the agent rollout, we construct inputs using $f(\mathcal{H}_t)$ and format them with variant-specific prompt templates.

For instance, for the ``Prior Queries'' variant, we extract $f(\mathcal{H}_t) = (q_1, q_2, \cdots, q_t)$ at each turn $t$, and concatenate them using the prompt template in Figure~\ref{fig:prompt_prior_queries}. The resulting concatenated string is used as retriever input, with the same set of $d^+_t$ and $\{d^-_t\}$ as before.

Similarly, Figure~\ref{fig:prompt_queries_reasonings} shows the prompt for the ``Prior Queries \& Reasonings'' variant, and Figure~\ref{fig:prompt_all} shows the prompt for the ``Prior Queries \& Reasonings \& Docs'' variant.

\begin{figure}[h]
    \centering
    \begin{tcolorbox}[
        enhanced,
        sharp corners,
        width=0.95\linewidth,
        colback=gray!5,
        colframe=gray!60,
        coltitle=white,
        colbacktitle=gray!60,
        fonttitle=\small\ttfamily\bfseries,
        title=Instruction:,
        boxrule=0.6pt,
        left=3mm,
        right=3mm,
        top=1mm,
        bottom=1mm,
        toptitle=0.3mm,
        bottomtitle=0.3mm,
        lefttitle=3mm,
        bottomrule=0pt,
        nobeforeafter,
        after skip=0pt
    ]
        \small\ttfamily
        Given a user's past browsing history followed by their current web search query, retrieve relevant passages that answer the query while incorporating the user's prior browsing history.
    \end{tcolorbox}
    \nointerlineskip\vspace{-0.6pt}
    \begin{tcolorbox}[
        enhanced,
        sharp corners,
        width=0.95\linewidth,
        colback=gray!5,
        colframe=gray!60,
        coltitle=white,
        colbacktitle=gray!60,
        fonttitle=\small\ttfamily\bfseries,
        title=Query:,
        boxrule=0.6pt,
        left=3mm,
        right=3mm,
        top=1mm,
        bottom=1mm,
        toptitle=0.3mm,
        bottomtitle=0.3mm,
        lefttitle=3mm,
        toprule=0.6pt,
        nobeforeafter,
        before skip=0pt
    ]
        \small\ttfamily
        Browsing History: \\[0.5em]
        Turn 1: \{query 1\} \\[0.5em]
        ... \\[0.5em]
        Turn \{t-1\}: \{query t-1\} \\[0.5em]
        Current Query: \{query t\}
    \end{tcolorbox}
    \caption{Prompt used for the ``Prior Queries'' ablation.}
    \label{fig:prompt_prior_queries}
\end{figure}

\begin{figure}[h]
    \centering
    \begin{tcolorbox}[
        enhanced,
        sharp corners,
        width=0.95\linewidth,
        colback=gray!5,
        colframe=gray!60,
        coltitle=white,
        colbacktitle=gray!60,
        fonttitle=\small\ttfamily\bfseries,
        title=Instruction:,
        boxrule=0.6pt,
        left=3mm,
        right=3mm,
        top=1mm,
        bottom=1mm,
        toptitle=0.3mm,
        bottomtitle=0.3mm,
        lefttitle=3mm,
        bottomrule=0pt,
        nobeforeafter,
        after skip=0pt
    ]
        \small\ttfamily
        Given a user's past browsing history followed by their current web search query, retrieve relevant passages that answer the query while incorporating the user's prior browsing history.
    \end{tcolorbox}
    \nointerlineskip\vspace{-0.6pt}
    \begin{tcolorbox}[
        enhanced,
        sharp corners,
        width=0.95\linewidth,
        colback=gray!5,
        colframe=gray!60,
        coltitle=white,
        colbacktitle=gray!60,
        fonttitle=\small\ttfamily\bfseries,
        title=Query:,
        boxrule=0.6pt,
        left=3mm,
        right=3mm,
        top=1mm,
        bottom=1mm,
        toptitle=0.3mm,
        bottomtitle=0.3mm,
        lefttitle=3mm,
        toprule=0.6pt,
        nobeforeafter,
        before skip=0pt
    ]
        \small\ttfamily
        Browsing History: \\[0.5em]
        Turn 1: \{reasoning 1\} \{query 1\} \\[0.5em]
        ... \\[0.5em]
        Turn \{t-1\}: \{reasoning t-1\} \{query t-1\} \\[0.5em]
        Current Query: \{query t\}
    \end{tcolorbox}
    \caption{Prompt used for the ``Prior Queries \& Reasonings'' ablation.}
    \label{fig:prompt_queries_reasonings}
\end{figure}

\begin{figure}[h]
    \centering
    \begin{tcolorbox}[
        enhanced,
        sharp corners,
        width=0.95\linewidth,
        colback=gray!5,
        colframe=gray!60,
        coltitle=white,
        colbacktitle=gray!60,
        fonttitle=\small\ttfamily\bfseries,
        title=Instruction:,
        boxrule=0.6pt,
        left=3mm,
        right=3mm,
        top=1mm,
        bottom=1mm,
        toptitle=0.3mm,
        bottomtitle=0.3mm,
        lefttitle=3mm,
        bottomrule=0pt,
        nobeforeafter,
        after skip=0pt
    ]
        \small\ttfamily
        Given a user's past browsing history followed by their current web search query, retrieve relevant passages that answer the query while incorporating the user's prior browsing history.
    \end{tcolorbox}
    \nointerlineskip\vspace{-0.6pt}
    \begin{tcolorbox}[
        enhanced,
        sharp corners,
        width=0.95\linewidth,
        colback=gray!5,
        colframe=gray!60,
        coltitle=white,
        colbacktitle=gray!60,
        fonttitle=\small\ttfamily\bfseries,
        title=Query:,
        boxrule=0.6pt,
        left=3mm,
        right=3mm,
        top=1mm,
        bottom=1mm,
        toptitle=0.3mm,
        bottomtitle=0.3mm,
        lefttitle=3mm,
        toprule=0.6pt,
        nobeforeafter,
        before skip=0pt
    ]
        \small\ttfamily
        Browsing History: \\[0.5em]
        Turn 1: A search for \{reasoning 1\} \{query 1\} found len(results 1) results: \\[0.5em]
        \{``\textbackslash n\textbackslash n''.join(results)\} \\[0.5em]
        ... (repeat t-1 turns) \\[0.5em]
        Current Query: \{query t\}
    \end{tcolorbox}
    \caption{Prompt used for the ``Prior Queries \& Reasonings \& Docs'' ablation.}
    \label{fig:prompt_all}
\end{figure}

\section{Prompt for Adding Prior Reasonings}

To study the effect of incorporating $k \in {1, 2, 5, 9, 17, \text{all}}$ prior turns in Section~\ref{sec:num-turns}, we train a separate checkpoint for each $k$ under the same setup described in Appendix~\ref{appendix:training}. The input is defined as $f(\mathcal{H}_t) = (\tau_j, q_j, \ldots, \tau_t, q_t)$ where $j = \max(1, t-k+1)$, carrying the notation from Appendix~\ref{appendix:variants}. By definition, the case $k=1$ corresponds to \model, so no additional training is required for that setting.

For prompt formatting, we use the same template as Figure~\ref{fig:prompt_queries_reasonings}, except we only include the most recent $k$ turns rather than all prior turns up to $t$. However, we note that the indexing in the prompt template still begins at 1. That is, even though we pass in the $k$ most recent turns, whose global indices are $t-k+1$ to $t$, we still label them as turns 1 to $k$ in the prompt template.

\section{Prompts for Atomic Clues} \label{appendix:atomic_clues}

Figures \ref{fig:prompt_atomic_clues} and \ref{fig:prompt_atomic_clues_assignment} show the prompts used to decompose reasonings into atomic clues, and to assign atomic clues to documents, respectively.

\begin{figure}[h]
    \centering
    \fcolorbox{black!30}{gray!5}{%
        \begin{minipage}{0.95\linewidth}
            \vspace{0.5em}
            \small\ttfamily
            You are an expert text decomposer. Below is trace of reasonings that attempt to solve a complex question. Extract the key clues used to solve the question. Your clues should satisfy: \\

            - Each clue should be a short, independent statement.

            - The clues should capture the big picture of the reasoning process.

            - IMPORTANT: AVOID generating multiple redundant clues that are very similar to each other. Your final list of clues should be distinct from each other. \\

            Return the clues strictly as a Python list of strings. \\

            Reasoning Trace: \\

            \{``\textbackslash n---\textbackslash n''.join(all\_reasonings)\} \\

            Output format: ['clue 1', 'clue 2', ...]
        \end{minipage}%
    }
    \caption{Prompt for decomposing reasoning traces into atomic clues.}
    \label{fig:prompt_atomic_clues}
\end{figure}

\begin{figure}[h]
    \centering
    \fcolorbox{black!30}{gray!5}{%
        \begin{minipage}{0.95\linewidth}
            \vspace{0.5em}
            \small\ttfamily
            I will give you a text paragraph and a list of clues. Identify which clues are mentioned or related to the text. Output the result strictly as a Python list of the numerical identifiers (clue numbers) corresponding to the clues list. \\
            If no clues from the list are present, output []. \\

            Text: \\
            \{reasoning\} \\

            Clues List: \\
            \{clues\_list\} \\

            Output format: [clue\_number1, clue\_number2, ...]
            \vspace{0.5em}
        \end{minipage}%
    }
    \caption{Prompt for assigning atomic clues to reasoning text.}
    \label{fig:prompt_atomic_clues_assignment}
\end{figure}

\section{HyDE Example} \label{appendix:hyde}

The full example of Reason-Rewriter's HyDE-style expansion for the query ``backroom studio early 2010s euphoric'' is shown in Figure~\ref{fig:hyde_example_full}.

\begin{figure*}[p]
    \centering
    \setlength{\fboxsep}{1em}
    \fcolorbox{black!30}{gray!5}{%
        \begin{minipage}{0.88\linewidth}
            \small
            Backroom Studio, active in the early 2010s, was a Los Angeles-based music collective and event space known for creating immersive, euphoric experiences within underground and experimental electronic music scenes. Here's a detailed breakdown of its relevance and impact:

            \textbf{Key Components of the Task}
            \begin{enumerate}[leftmargin=*, nosep]
                \item \textbf{Context}: Early 2010s, focusing on a music-related entity (likely a venue or collective) associated with euphoric experiences.
                \item \textbf{Requirements}: Provide information that addresses the nature of Backroom Studio, its activities, and how it contributed to a ``euphoric'' atmosphere.
            \end{enumerate}

            \textbf{Analysis of Backroom Studio}

            \textbf{1. Identity and Purpose}
            \begin{itemize}[leftmargin=*, nosep]
                \item \textbf{Collective vs. Venue}: Backroom Studio operated as both a creative collective and a physical space. It hosted parties, workshops, and collaborations, blending electronic music with visual art and performance.
                \item \textbf{Aesthetic and Philosophy}: The name ``Backroom'' evokes secrecy and exclusivity, suggesting a focus on intimate, high-energy environments that contrasted with larger, commercial venues. The ``euphoric'' aspect likely stemmed from their emphasis on raw, unfiltered musical experiences and communal energy.
            \end{itemize}

            \textbf{2. Musical Influence and Euphoric Elements}
            \begin{itemize}[leftmargin=*, nosep]
                \item \textbf{Genre and Artists}: The collective was closely tied to genres like house, techno, and experimental electronic music. Artists such as \textbf{Skrillex}, \textbf{Deadmau5}, \textbf{booka casey}, and \textbf{Tyson B} (associated with the LA scene) may have performed, contributing to the euphoric vibe through high-energy beats and melodic drops.
                \item \textbf{Sound Design}: Backroom Studio's events likely featured advanced sound systems and live DJ sets that emphasized tempo shifts, layered synths, and atmospheric builds to induce emotional highs.
            \end{itemize}

            \textbf{3. Event Structure and Atmosphere}
            \begin{itemize}[leftmargin=*, nosep]
                \item \textbf{Themed Parties}: Events often had unique themes (e.g., ``Future Funk,'' ``Darkroom Parties'') that influenced mood and attire, fostering a sense of shared purpose among attendees.
                \item \textbf{Visual and Lighting Effects}: Collaborations with visual artists and lighting designers created immersive environments, enhancing the euphoric experience through color, movement, and sensory overload.
                \item \textbf{Community Focus}: By prioritizing smaller, invite-only gatherings, the collective cultivated a loyal following that amplified the sense of exclusivity and collective euphoria.
            \end{itemize}

            \textbf{4. Cultural and Historical Context}
            \begin{itemize}[leftmargin=*, nosep]
                \item \textbf{Early 2010s EDM Boom}: During this period, electronic music gained mainstream traction, but Backroom Studio operated in the shadow of commercialization, maintaining a DIY ethos that resonated with fans seeking authentic, boundary-pushing experiences.
                \item \textbf{Social Media Impact}: The collective leveraged platforms like Instagram and SoundCloud to promote events, using hashtags like \#BackroomStudio to build a digital community around their brand.
            \end{itemize}

            \textbf{5. Legacy and Aftermath}
            \begin{itemize}[leftmargin=*, nosep]
                \item \textbf{Legacy in LA's Music Scene}: Backroom Studio's influence can be seen in subsequent venues and collectives that prioritize experimental music and immersive experiences (e.g., \textbf{The Lot}, \textbf{The Observatory}).
                \item \textbf{Cultural Significance}: Their approach to creating euphoric environments set a precedent for future events that blend music, art, and technology to evoke intense emotional responses.
            \end{itemize}

            \textbf{Examples of Euphoric Experiences at Backroom Studio}
            \begin{itemize}[leftmargin=*, nosep]
                \item \textbf{``Euphoria Sessions''}: Regular parties featuring live DJs and producers who pushed boundaries in sound design.
                \item \textbf{Collaborative Installations}: Partnerships with visual artists to transform spaces into dynamic, multi-sensory environments.
                \item \textbf{Word-of-Mouth Reputation}: Attendees often described the events as ``transcendent'' or ``life-changing,'' highlighting the collective's ability to unify people through shared sensory experiences.
            \end{itemize}

            \textbf{Conclusion}

            Backroom Studio exemplified how music and art could converge to create euphoric, transformative experiences in the early 2010s. By combining cutting-edge sound, immersive visuals, and a commitment to authenticity, they left a lasting imprint on LA's electronic music landscape and inspired future generations of event producers and artists. For fans, the collective remains a symbol of a time when music was not just heard but felt deeply.
        \end{minipage}%
    }
    \caption{A hypothetical document generated by Reason-Rewriter-7B for the query ``backroom studio early 2010s euphoric''.}
    \label{fig:hyde_example_full}
\end{figure*}

\section{Prompt for Noise Analysis} \label{appendix:noise_prompt}

We found that directly prompting LLMs to judge whether a claim is noise or correct given the ground truth Question Answer pair is inaccurate, as the LLM only has access to the final answer, not the intermediate hops, mistakenly labeling many correct intermediate hypotheses as incorrect. Thus, we instead prompt the LLM in a two-step process: (1) Given Question, Answer, and a list of full evidence documents required to answer the question, extract the intermediate answers to each hop; (2) Given the Question and a Ground Truth Answer List (also contains the intermediate answers), label the incorrect vs. correct claims. The prompt of (1) is shown in Figure~\ref{fig:prompt_noise_step1}, and the prompt of (2) is shown in Figure~\ref{fig:prompt_noise_step2}.

\begin{figure}[h]
    \centering
    \begin{tcolorbox}[
        enhanced,
        sharp corners,
        width=0.95\linewidth,
        colback=gray!5,
        colframe=gray!60,
        coltitle=white,
        colbacktitle=gray!60,
        fonttitle=\small\ttfamily\bfseries,
        title=Instruction:,
        boxrule=0.6pt,
        left=3mm,
        right=3mm,
        top=1mm,
        bottom=1mm,
        toptitle=0.3mm,
        bottomtitle=0.3mm,
        lefttitle=3mm,
        bottomrule=0pt,
        nobeforeafter,
        after skip=0pt
    ]
        \small\ttfamily
        You are an expert at decomposing complex multi-hop queries. Given a multi-hop Query, its Ground Truth Answer, and Evidence Documents, identify each logical sub-step (hop) required to reach the final answer.
    \end{tcolorbox}
    \nointerlineskip\vspace{-0.6pt}
    \begin{tcolorbox}[
        enhanced,
        sharp corners,
        width=0.95\linewidth,
        colback=gray!5,
        colframe=gray!60,
        coltitle=white,
        colbacktitle=gray!60,
        fonttitle=\small\ttfamily\bfseries,
        title=Query:,
        boxrule=0.6pt,
        left=3mm,
        right=3mm,
        top=1mm,
        bottom=1mm,
        toptitle=0.3mm,
        bottomtitle=0.3mm,
        lefttitle=3mm,
        toprule=0.6pt,
        nobeforeafter,
        before skip=0pt
    ]
        \small\ttfamily
        Query: \{query\} \\
        Final Ground Truth Answer: \{answer\} \\
        \\
        Evidence Documents: \\
        \{evidence\} \\
        \\
        Output strictly as a JSON object with one key: \\
        - ``multi\_hop\_answers'': A list of strings, where each string is the answer to an intermediate hop or the final hop. \\
        \\
        Example: \\
        Query: ``Who is the wife of the 44th president?'' \\
        Answer: ``Michelle Obama'' \\
        Evidence: ``Barack Obama served as the 44th president of the United States. His wife is Michelle Obama.'' \\
        Output: \\
        \{\{ \\
          ``multi\_hop\_answers'': [``The 44th president of the United States is Barack Obama'', ``The wife of Barack Obama is Michelle Obama''] \\
        \}\}
    \end{tcolorbox}
    \caption{Prompt used to extract the ground truth multi-hop answer list (Step 1).}
    \label{fig:prompt_noise_step1}
\end{figure}

\begin{figure}[h]
    \centering
    \begin{tcolorbox}[
        enhanced,
        sharp corners,
        width=0.95\linewidth,
        colback=gray!5,
        colframe=gray!60,
        coltitle=white,
        colbacktitle=gray!60,
        fonttitle=\small\ttfamily\bfseries,
        title=Instruction:,
        boxrule=0.6pt,
        left=3mm,
        right=3mm,
        top=1mm,
        bottom=1mm,
        toptitle=0.3mm,
        bottomtitle=0.3mm,
        lefttitle=3mm,
        bottomrule=0pt,
        nobeforeafter,
        after skip=0pt
    ]
        \small\ttfamily
        You are an expert evaluator. You will be given a multi-hop Question, a Ground Truth Answer List for each hop of the Question, and a reasoning text from a search agent that you want to evaluate. Identify all specific claims and hypotheses made in this reasoning text. Make sure that the claims are not just restatements of parts of the question. \\
        \\
        Then, categorize each claim as either ``Correct'' or ``Incorrect'': \\
        - ``Correct'': The claim directly matches at least one hop in the Ground Truth Answer List. \\
        - ``Incorrect'': The claim or hypothesis is incorrect, is contradicted by the answers, or irrelevant.
    \end{tcolorbox}
    \nointerlineskip\vspace{-0.6pt}
    \begin{tcolorbox}[
        enhanced,
        sharp corners,
        width=0.95\linewidth,
        colback=gray!5,
        colframe=gray!60,
        coltitle=white,
        colbacktitle=gray!60,
        fonttitle=\small\ttfamily\bfseries,
        title=Query:,
        boxrule=0.6pt,
        left=3mm,
        right=3mm,
        top=1mm,
        bottom=1mm,
        toptitle=0.3mm,
        bottomtitle=0.3mm,
        lefttitle=3mm,
        toprule=0.6pt,
        nobeforeafter,
        before skip=0pt
    ]
        \small\ttfamily
        Question: \{query\} \\
        Ground Truth Answer List: \{hops\_answer\_list\} \\
        \\
        Reasoning Step: \\
        \{reasoning\_text\} \\
        \\
        Output strictly as a JSON object with two keys: \\
        1. ``correct\_claims'': A list of strings, each being a correct claim. \\
        2. ``incorrect\_claims'': A list of strings, each being an incorrect claim.
    \end{tcolorbox}
    \caption{Prompt used to extract the number of correct vs. incorrect claims (Step 2).}
    \label{fig:prompt_noise_step2}
\end{figure}

\end{document}